\documentclass[sigconf]{acmart}
\usepackage{booktabs} % For formal tables
\usepackage{url}
\usepackage{amsmath}
\usepackage{amssymb}
\usepackage{dirtytalk}
\usepackage{graphicx}
\usepackage{hyperref}
\usepackage{wrapfig}
\renewcommand{\vec}[1]{\mathbf{#1}}

\usepackage{listings}
\usepackage{color}

\definecolor{dkgreen}{rgb}{0,0.6,0}
\definecolor{gray}{rgb}{0.5,0.5,0.5}
\definecolor{mauve}{rgb}{0.58,0,0.82}

\begin{document}

\copyrightyear{2018}
\acmYear{2018}
\setcopyright{acmlicensed}
\acmConference[ICMLSC 2018]{ICMLSC 2018, The 2nd International Conference on Machine Learning and Soft Computing}{February 2--4, 2018}{Phu Quoc Island, Viet Nam}
\acmBooktitle{ICMLSC 2018: ICMLSC 2018, The 2nd International Conference on Machine Learning and Soft Computing, February 2--4, 2018, Phu Quoc Island, Viet Nam}
\acmPrice{15.00}
\acmDOI{10.1145/3184066.3184080}
\acmISBN{978-1-4503-6336-5/18/02}

\title[On Breast Cancer Detection]{On Breast Cancer Detection: An Application of Machine Learning Algorithms on the Wisconsin Diagnostic Dataset}

\author{Abien Fred M. Agarap}
\email{abienfred.agarap@gmail.com}

\begin{abstract}
This paper presents a comparison of six machine learning (ML) algorithms: GRU-SVM\cite{agarap2017neural}, Linear Regression, Multilayer Perceptron (MLP), Nearest Neighbor (NN) search, Softmax Regression, and Support Vector Machine (SVM) on the Wisconsin Diagnostic Breast Cancer (WDBC) dataset\cite{wolberg1992breast} by measuring their classification test accuracy, and their sensitivity and specificity values. The said dataset consists of features which were computed from digitized images of FNA tests on a breast mass\cite{wolberg1992breast}. For the implementation of the ML algorithms, the dataset was partitioned in the following fashion: 70\% for training phase, and 30\% for the testing phase. The hyper-parameters used for all the classifiers were manually assigned. Results show that all the presented ML algorithms performed well (all exceeded 90\% test accuracy) on the classification task. The MLP algorithm stands out among the implemented algorithms with a test accuracy of $\approx$99.04\%.
\end{abstract}




\maketitle
\section{Introduction}
Breast cancer is one of the most common cancer along with lung and bronchus cancer, prostate cancer, colon cancer, and pancreatic cancer among others\cite{nci2017}. Representing 15\% of all new cancer cases in the United States alone\cite{surveillance}, it is a topic of research with great value.\\
\indent	The utilization of data science and machine learning approaches in medical fields proves to be prolific as such approaches may be considered of great assistance in the decision making process of medical practitioners. With an unfortunate increasing trend of breast cancer cases\cite{surveillance}, comes also a big deal of data which is of significant use in furthering clinical and medical research, and much more to the application of data science and machine learning in the aforementioned domain.\\
\indent	Prior studies have seen the importance of the same research topic\cite{salama2012breast, zafiropoulos2006support}, where they proposed the use of machine learning (ML) algorithms for the classification of breast cancer using the Wisconsin Diagnostic Breast Cancer (WDBC) dataset\cite{wolberg1992breast}, and eventually had significant results.\\
\indent	This paper presents yet another study on the said topic, but with the introduction of our recently-proposed GRU-SVM model\cite{agarap2017neural}. The said ML algorithm combines a type of recurrent neural network (RNN), the gated recurrent unit (GRU)\cite{Cho} with the support vector machine (SVM)\cite{Cortes}. Along with the GRU-SVM model, a number of ML algorithms is presented in Section \ref{ml-algorithms}, which were all applied on breast cancer classification with the aid of WDBC\cite{wolberg1992breast}.

\section{Methodology}

\subsection{Machine Intelligence Library}
Google TensorFlow\cite{tensorflow2015-whitepaper} was used to implement the machine learning algorithms in this study, with the aid of other scientific computing libraries: matplotlib\cite{Hunter:2007}, numpy\cite{walt2011numpy}, and scikit-learn\cite{scikit-learn}.

\subsection{The Dataset}
The machine learning algorithms were trained to detect breast cancer using the Wisconsin Diagnostic Breast Cancer (WDBC) dataset\cite{wolberg1992breast}. According to \cite{wolberg1992breast}, the dataset consists of features which were computed from a digitized image of a fine needle aspirate (FNA) of a breast mass. The said features describe the characteristics of the cell nuclei found in the image\cite{wolberg1992breast}.
\begin{figure}[!htb]
\minipage{0.45\textwidth}
\centering
	\includegraphics[width=\linewidth]{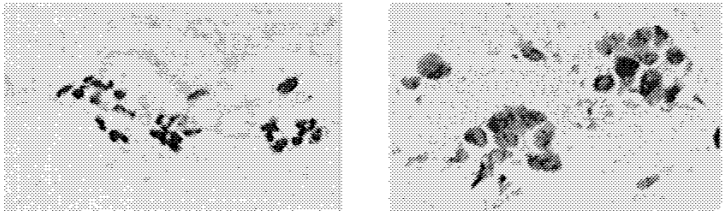}
	\caption{Image from \cite{wolberg1992breast} as cited by \cite{zafiropoulos2006support}. Digitized images of FNA: (a) Benign, (b) Malignant.}
	\label{training-accuracy}
\endminipage\hfill
\end{figure}

\indent	There are 569 data points in the dataset: 212 -- Malignant, 357 -- Benign. Accordingly, the dataset features are as follows: (1) radius, (2) texture, (3) perimeter, (4) area, (5) smoothness, (6) compactness, (7) concavity, (8) concave points, (9) symmetry, and (10) fractal dimension. With each feature having three information\cite{wolberg1992breast}: (1) mean, (2) standard error, and (3) ``worst'' or largest (mean of the three largest values) computed. Thus, having a total of 30 dataset features.

\subsection{Dataset Preprocessing}

To avoid inappropriate assignment of relevance, the dataset was standardized using Eq. \ref{student-t-stat}.

\begin{equation}\label{student-t-stat}
z=\dfrac{X - \mu}{\sigma}
\end{equation}

where $X$ is the feature to be standardized, $\mu$ is the mean value of the feature, and $\sigma$ is the standard deviation of the feature. The standardization was implemented using \texttt{StandardScaler().fit\_transform()} of \texttt{scikit-learn}\cite{scikit-learn}.

\subsection{Machine Learning (ML) Algorithms}\label{ml-algorithms}

This section presents the machine learning (ML) algorithms used in the study. The Stochastic Gradient Descent (SGD) learning algorithm was used for all the ML algorithms presented in this section except for GRU-SVM, Nearest Neighbor search, and Support Vector Machine. The code implementations may be found online at https://github.com/AFAgarap/wisconsin-breast-cancer.

\subsubsection{GRU-SVM}
We proposed a neural network architecture\cite{agarap2017neural} combining the gated recurrent unit (GRU) variant of recurrent neural network (RNN) and the support vector machine (SVM), for the purpose of binary classification.
\begin{equation}\label{z-gate}
z	=	\sigma(\vec{W}_{z} \cdot [h_{t - 1}, x_{t}])
\end{equation}
\begin{equation}\label{r-gate}
r	=	\sigma(\vec{W}_{r} \cdot [h_{t - 1}, x_{t}])
\end{equation}
\begin{equation}\label{candidate-value}
\tilde{h}_{t}	=	tanh(\vec{W} \cdot [r_{t} * h_{t - 1}, x_{t}])
\end{equation}
\begin{equation}\label{new-value}
h_{t}	=	(1 - z_{t}) * h_{t - 1} + z_{t} * \tilde{h}_{t}
\end{equation}

where $z$ and $r$ are the \textit{update gate} and \textit{reset gate} of a GRU-RNN respectively, $\tilde{h}_{t}$ is the candidate value, and $h_{t}$ is the new RNN cell state value\cite{Cho}. In turn, the $h_{t}$ is used as the predictor variable $\vec{x}$ in the L2-SVM predictor function (given by $sign(\vec{w}\vec{x} + b)$) of the network instead of the conventional Softmax classifier. \\
\indent The learning parameter $\vec{W}$ of the GRU-RNN is learned by the L2-SVM using the loss function given by Eq. \ref{l2-svm}. The computed loss is then minimized through Adam\cite{Kingma} optimization. The same optimization algorithm was used for Softmax Regression (Section \ref{softmax-regression}) and SVM (Section \ref{svm}). Then, the decision function $f(x) = sign(\vec{w}\vec{x} + b)$ produces a vector of scores for each cancer diagnosis: -1 for benign, and +1 for malignant. In order to get the predicted labels $y$ for a given data $\vec{x}$, the $argmax$ function is used (see Eq. \ref{argmax}).
\begin{align}\label{argmax}
\vec{y'}	&=	argmax\big(sign(\vec{w}\vec{x} + b)\big)
\end{align}

\indent	The $argmax$ function shall return the indices of the highest scores across the vector of predicted classes $sign(\vec{w}\vec{x} + b)$.

\subsubsection{Linear Regression}\label{linear-regression}
Despite an algorithm for regression problem, linear regression (see Eq. \ref{linear-regression-equation}) was used as a classifier for this study. This was done by applying a threshold for the output of Eq. \ref{linear-regression-equation}, i.e. subjecting the value of the regressand to Eq. \ref{threshold}.
\begin{align}\label{linear-regression-equation}
h_{\theta}(x)	&= \sum_{i=0}^{n} \theta_{i} \cdot x_{i} + b
\end{align}
\begin{align}\label{threshold}
f\big(h_{\theta}(x)\big)	=	\begin{cases}
									1	&	h_{\theta}(x) \geq 0.5	\\
									0	&	h_{\theta}(x) < 0.5
								\end{cases}
\end{align}

To measure the loss of the model, the mean squared error (MSE) was used (see Eq. \ref{mse}).
\begin{align}\label{mse}
L(y, \theta, x)	&=	\dfrac{1}{N} \sum_{i = 0}^{N} \big(y_{i} - (\theta_{i} \cdot x_{i} + b)\big)^{2}
\end{align}

where $y$ represents the actual class, and $(\vec{\theta} \cdot \vec{x} + b)$ represents the predicted class. This loss is minimized using the SGD algorithm, which learns the parameters $\vec{\theta}$ of Eq. \ref{linear-regression-equation}. The same method of loss minimization was used for MLP and Softmax Regression.

\subsubsection{Multilayer Perceptron}
The perceptron model was developed by Rosenblatt (1958)\cite{rosenblatt1958perceptron} based on the neuron model by McCulloch \& Pitts (1943)\cite{mcculloch1943logical}. The multilayer perceptron (MLP)\cite{bishop1995neural} consists of hidden layers (composed by a number of perceptrons) that enable the approximation of any functions, that is, through activation functions such as $tanh$ or \textit{sigmoid} $\sigma$.
\begin{align}
h_{\theta}(x)	&=	\sum_{i = 0}^{n} \theta_{i} x_{i} + b
\end{align}
\begin{align}\label{relu}
f\big(h_{\theta}(x)\big)	&= \vec{h_{\theta}(x)}^{+} = max(0, \vec{h_{\theta}(x)})
\end{align}

For this study, the activation function used for MLP was ReLU\cite{hahnloser2000digital} (see Eq. \ref{relu}), while there were three hidden layers that each consists of 500 nodes (500-500-500 architecture). As for the loss, it was computed using the cross entropy function (see Eq. \ref{cross-entropy}).

\subsubsection{Nearest Neighbor}
This is a form of an optimization problem that seeks to find the closest point $p_{i} \in \vec{p}$ to a query point $q_{i} \in \vec{q}$. In this study, both the L1 (Manhattan, see Eq. \ref{taxicab-norm}) and L2 (Euclidean, see Eq. \ref{euclidean-norm}) norm were used to measure the distance between $\vec{p}$ and $\vec{q}$.
\begin{equation}\label{taxicab-norm}
        \begin{gathered}
                \|\vec{p} - \vec{q}\|_{1} = \sum_{i=1}^{n} |p_{i} - q_{i}|
        \end{gathered}
\end{equation}
\begin{equation}\label{euclidean-norm}
        \begin{gathered}
                \|\vec{p} - \vec{q}\|_{2} = \sqrt{\sum_{i=1}^{n} (p_{i} - q_{i})^{2}}
        \end{gathered}
\end{equation}

The code implementation was based on the work of Damien (2017)\cite{aymericdamien} in GitHub. A learning algorithm such as SGD and Adam\cite{Kingma} is not applicable to Nearest Neighbor search, as it is practically a geometric approach for classification.

\subsubsection{Softmax Regression}\label{softmax-regression}
This is a classification model generalizing logistic regression to multinomial problems. But unlike linear regression (Section \ref{linear-regression}) that produces raw scores for the classes, softmax regression produces a probability distribution for the classes. This is accomplished using the Softmax function (see Eq. \ref{softmax}).
\begin{align}\label{softmax}
P(\vec{\hat{y}}\ |\ \vec{x})	&= \dfrac{e^{\hat{y}_{i}}}{\sum_{i=0}^{n} e^{\hat{y}_{i}}}
\end{align}

\begin{align}\label{cross-entropy}
L(\vec{y}, \vec{\hat{y}})	&=	-\sum_{i = 0}^{n} y_{i} \cdot log\big(\hat{y}_{i}\big)
\end{align}

The loss is measured by using the cross entropy function (see Eq. \ref{cross-entropy}), where $\vec{y}$ represents the actual class, and $\vec{\hat{y}}$ represents the predicted class.

\subsubsection{Support Vector Machine}\label{svm}
Developed by Vapnik\cite{Cortes}, the support vector machine (SVM) was primarily intended for binary classification. Its main objective is to determine the optimal hyperplane $f(w, x) = \vec{w} \cdot \vec{x} + b$ separating two classes in a given dataset having input features $\vec{x} \in \mathbb{R}^{p}$, and labels $y \in \{-1, +1\}$.

SVM learns by solving the following constrained optimization problem:
\begin{equation} \label{constrained-l1}
min \dfrac{1}{p}\vec{w}^{T}\vec{w} + C \sum_{i = 1}^{p} \xi_i
\end{equation}
\begin{align}
s.t\ y_{i}'(\vec{w} \cdot \vec{x} + b) \geq 1 - \xi_i\\
\xi_i \geq 0, i = 1, ..., p
\end{align}

where $\vec{w}^{T} \vec{w}$ is the Manhattan norm, $\xi$ is a cost function, and $C$ is the penalty parameter (may be an arbitrary value or a selected value using hyper-parameter tuning). Its corresponding unconstrained optimization problem is the following:
\begin{equation} \label{l1-svm}
min \dfrac{1}{p}\vec{w}^{T}\vec{w} + C \sum_{i = 1}^{p} max\big(0, 1 - y_{i}'(w_{i}x_{i}+b)\big)
\end{equation}

\indent where $\vec{w}\vec{x} + b$ is the predictor function. The objective of Eq. \ref{l1-svm} is known as the primal form problem of L1-SVM, with the standard hinge loss. The problem with L1-SVM is the fact that it is not differentiable\cite{Tang}, as opposed to its variation, the L2-SVM:
\begin{equation}\label{l2-svm}
min \dfrac{1}{p}\|\vec{w}\|_{2}^{2} + C \sum_{i = 1}^{p} max\big(0, 1 - y_{i}'(w_{i}x_{i}+b)\big)^{2}
\end{equation}

The L2-SVM is differentiable and provides more stable results than its L1 counterpart\cite{Tang}.

\subsection{Data Analysis}
There were two phases of experiment for this study: (1) training phase, and (2) test phase. The dataset was partitioned by 70\% (training phase) / 30\% (testing phase). The parameters considered in the experiments were as follows: (1) Test Accuracy, (2) Epochs, (3) Number of data points, (4) False Positive Rate (FPR), (5) False Negative Rate (FNR), (6) True Positive Rate (TPR), and (7) True Negative Rate (TNR).

% \begin{align}\label{TPR}
% TPR	&=	\dfrac{True\ Positive}{True\ Positive + False\ Negative}
% \end{align}
% \begin{align}\label{TNR}
% TNR	&=	\dfrac{True\ Negative}{True\ Negative + False\ Positive}
% \end{align}
% \begin{align}\label{FPR}
% FPR	&=	1 - TNR
% \end{align}
% \begin{align}\label{FNR}
% FNR	&=	1 - TPR
% \end{align}

\section{Results and Discussion}
All experiments in this study were conducted on a laptop computer with Intel Core(TM) i5-6300HQ CPU @ 2.30GHz x 4, 16GB of DDR3 RAM, and NVIDIA GeForce GTX 960M 4GB DDR5 GPU. Table \ref{table: hyperparameters} shows the manually-assigned hyper-parameters used for the ML algorithms. Table \ref{table: summary-results} summarizes the experiment results. In addition to the reported results, the result from \cite{zafiropoulos2006support} was put into comparison.\\
\begin{table*}
\centering
\caption{Hyper-parameters used for the ML algorithms.}
		\begin{tabular}{ccccccc}
		\toprule
		Hyper-parameters & GRU-SVM & Linear Regression & MLP & Nearest Neighbor & Softmax Regression & SVM \\
		\midrule
		Batch Size & 128 & 128 & 128 & N/A & 128 & 128 \\
		Cell Size & 128 & N/A & [500, 500, 500] & N/A & N/A & N/A \\
		Dropout Rate & 0.5 & N/A & None & N/A & N/A & N/A\\
		Epochs & 3000 & 3000 & 3000 & 1 & 3000 & 3000\\
		Learning Rate & 1e-3 & 1e-3 & 1e-2 & N/A & 1e-3 & 1e-3\\
		Norm & L2 & N/A & N/A & L1, L2 & N/A & L2\\
		SVM C & 5 & N/A & N/A & N/A & N/A & 5\\
		\bottomrule
		\end{tabular}\\
		\label{table: hyperparameters}
\end{table*}
\begin{table*}
\centering
\caption{Summary of experiment results on the ML algorithms.}
		\begin{tabular}{cccccccc}
		\toprule
		Parameter & GRU-SVM & Linear Regression & MLP & L1-NN & L2-NN & Softmax Regression & SVM \\
		\midrule
		Accuracy & 93.75\% & 96.09375\% & 99.038449585420729\% & 93.567252\% & 94.736844\% & 97.65625\% & 96.09375\% \\
		Data points  & 384000 & 384000 & 512896 & 171 & 171 & 384000 & 384000 \\
		Epochs & 3000 & 3000 & 3000 & 1 & 1 & 3000 & 3000\\
		FPR & 16.666667\% & 10.204082\% & 1.267042\% & 6.25\% & 9.375\% & 5.769231\% & 6.382979\% \\
		FNR & 0 & 0 & 0.786157\% & 6.542056\% & 2.803738\%& 0 & 2.469136\% \\
		TPR & 100\% & 100\% & 99.213843\% & 93.457944\% & 97.196262\% & 100\% & 97.530864\% \\
		TNR & 83.333333\% & 89.795918\% & 98.732958\% & 93.75\% & 90.625\% & 94.230769\% & 93.617021\% \\
		\bottomrule
		\end{tabular}\\
		\label{table: summary-results}
\end{table*}
% \indent	First, \cite{salama2012breast} implemented five ML algorithms for classification on WDBC. Namely, (1) Naive Bayes (NB), (2) MLP, (3) SVM, (4) K-Nearest Neighbor (KNN), and (5) Decision Tree J48\cite{quinlan1993c4}. The single classifier algorithms they used were implemented using WEKA\cite{witten2016data}, and their hyper-parameters and architecture (for MLP) were not specified. Their results have shown the following test accuracies: (1) NB had 92.9701\%, (2) MLP had 96.6608\%, (3) SVM had 97.71533\%, (4) KNN had 95.9575\%, and (5) Decision Tree J48 had 93.1459\%. However, they did not indicate the specifics of their dataset partition, i.e. how many percent was for the training data, and how many percent was for the testing data.\\
\indent	\cite{zafiropoulos2006support} implemented the SVM with Gaussian Radial Basis Function (RBF) as its kernel for classification on WDBC. Their experiment revealed that their SVM had its highest test accuracy of 89.28\% with its free parameter $\sigma = 0.6$. However, their experiment was based on a 60/40 partition (training/testing respectively).
% \begin{align}\label{gaussian-rbf}
% 	K(x, x')	&=	exp\bigg(-\dfrac{\|x - x'\|^{2}}{2\sigma^{2}}\bigg)
% \end{align}
Hence, we would not be able to draw a fair comparison between the current study and \cite{zafiropoulos2006support}. Comparing the results of this study on an intuitive sense may perhaps be close to a fair comparison, recalling that the partition done in this study was 70/30.\\
\indent	With a test accuracy of $\approx$96.09\%, the L2-SVM in this study bares superiority against the findings of \cite{zafiropoulos2006support} (SVM with Gaussian RBF, having a test accuracy of 89.28\%). But then again, it was based on a higher training data of 10\% (70\% vs 60\%).\\
\begin{figure}[!htb]
\minipage{0.5\textwidth}\centering
	\includegraphics[width=\linewidth]{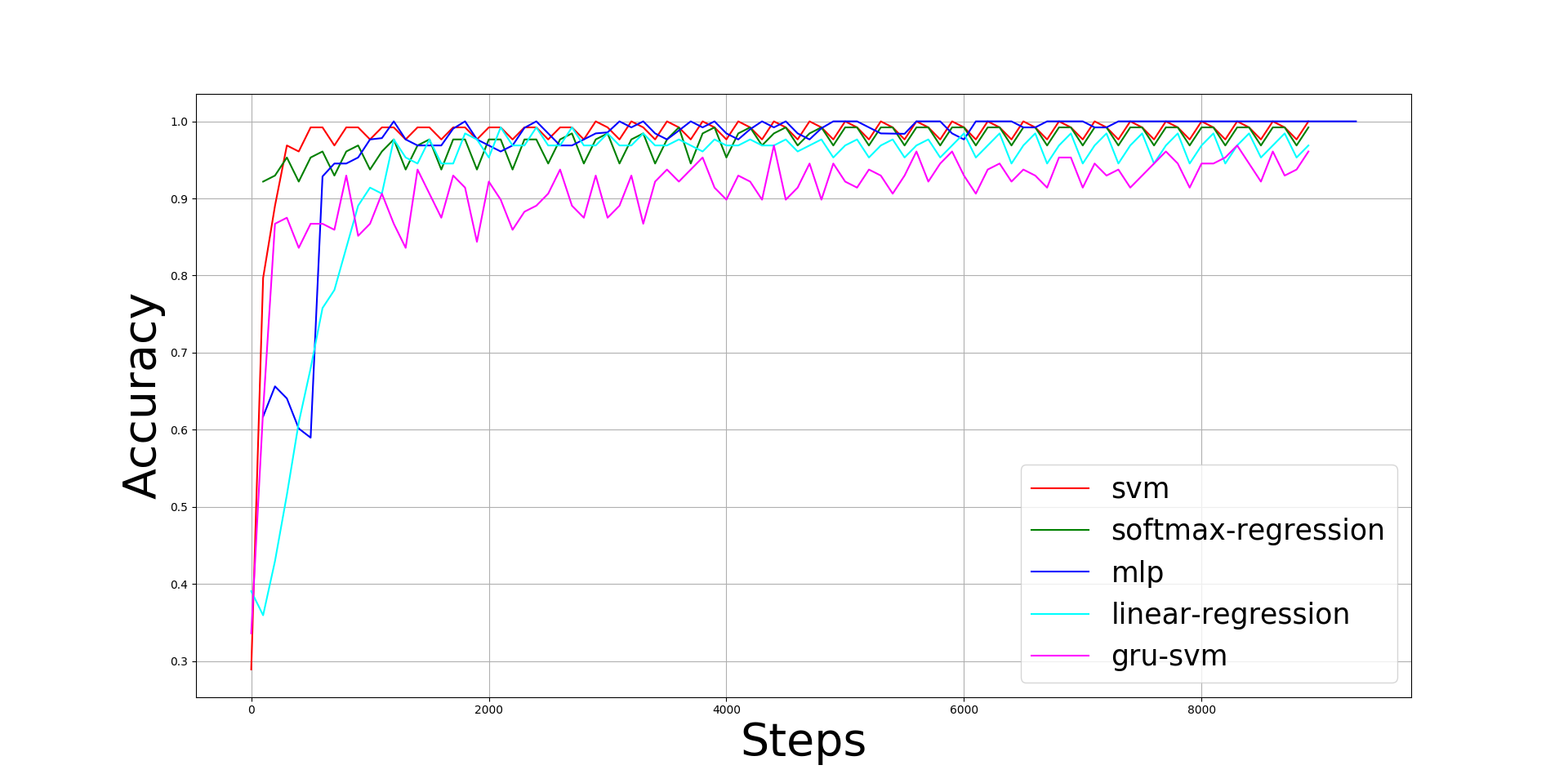}
	\caption{Plotted using \texttt{matplotlib}\cite{Hunter:2007}. Training accuracy of the ML algorithms on breast cancer detection using WDBC.}
	\label{training-accuracy}
\endminipage\hfill
\end{figure}
\begin{figure}[!htb]
\minipage{0.5\textwidth}\centering
	\includegraphics[width=\linewidth]{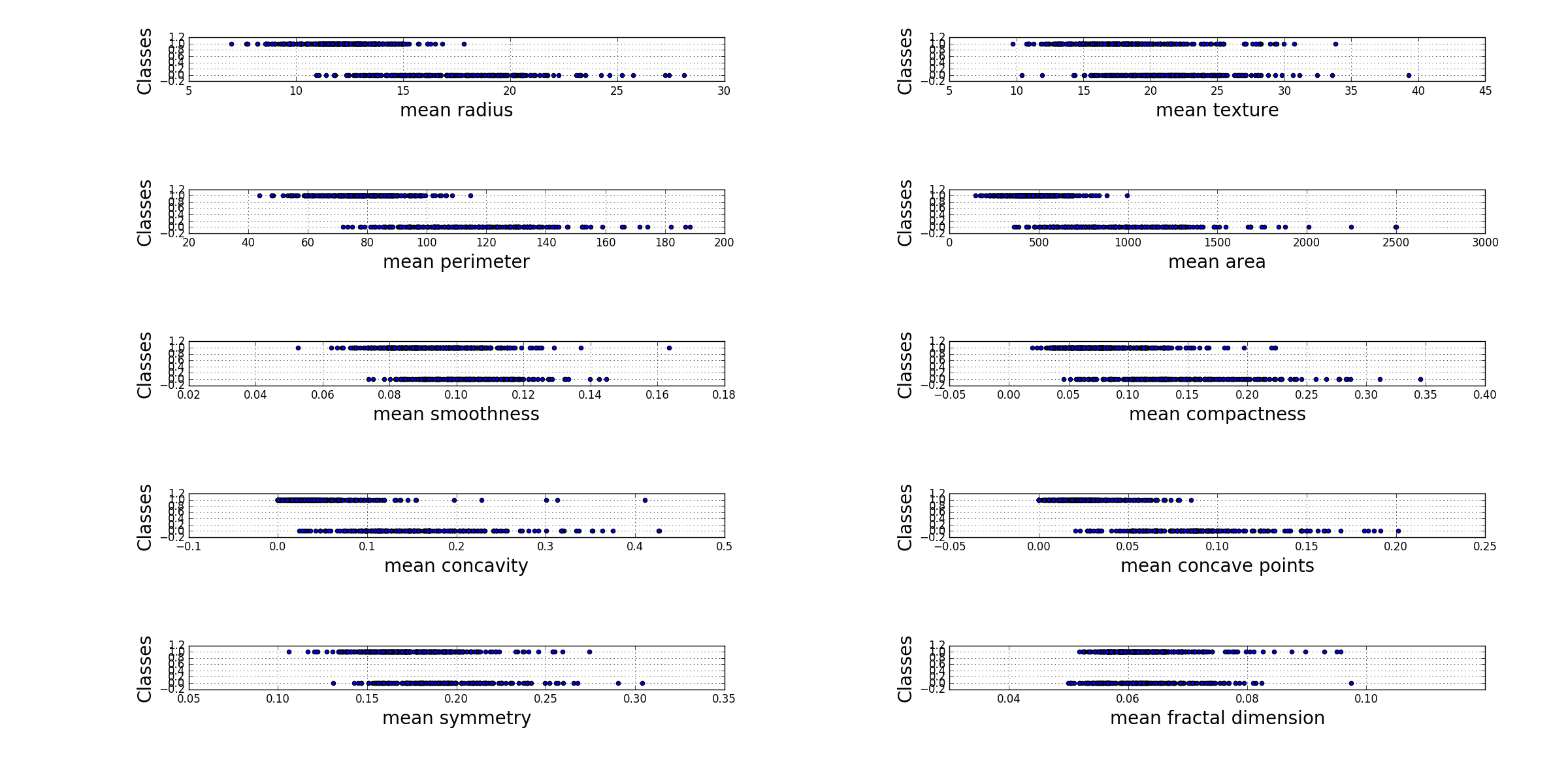}
	\caption{Plotted using \texttt{matplotlib}\cite{Hunter:2007}. Scatter plot of \textit{mean} features ($x_{0} - x_{9}$) in the WDBC.}
	\label{scatter-plot-mean}
\endminipage\hfill
\end{figure}
\begin{figure}[!htb]
\minipage{0.5\textwidth}\centering
	\includegraphics[width=\linewidth]{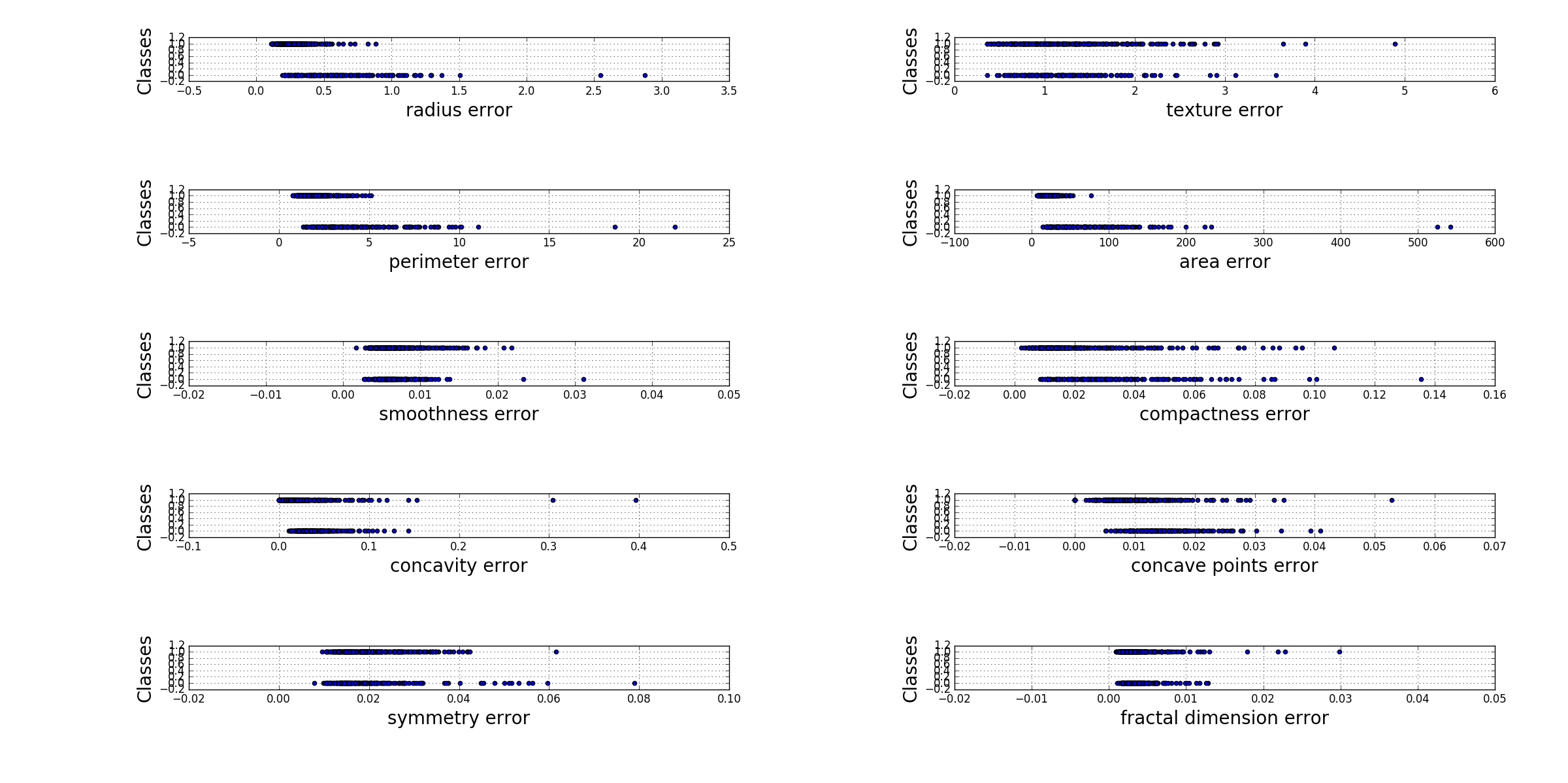}
	\caption{Plotted using \texttt{matplotlib}\cite{Hunter:2007}. Scatter plot of \textit{error} features ($x_{10} - x_{19}$) in the WDBC.}
	\label{scatter-plot-error}
\endminipage\hfill
\end{figure}
\begin{figure}[!htb]
\minipage{0.5\textwidth}\centering
	\includegraphics[width=\linewidth]{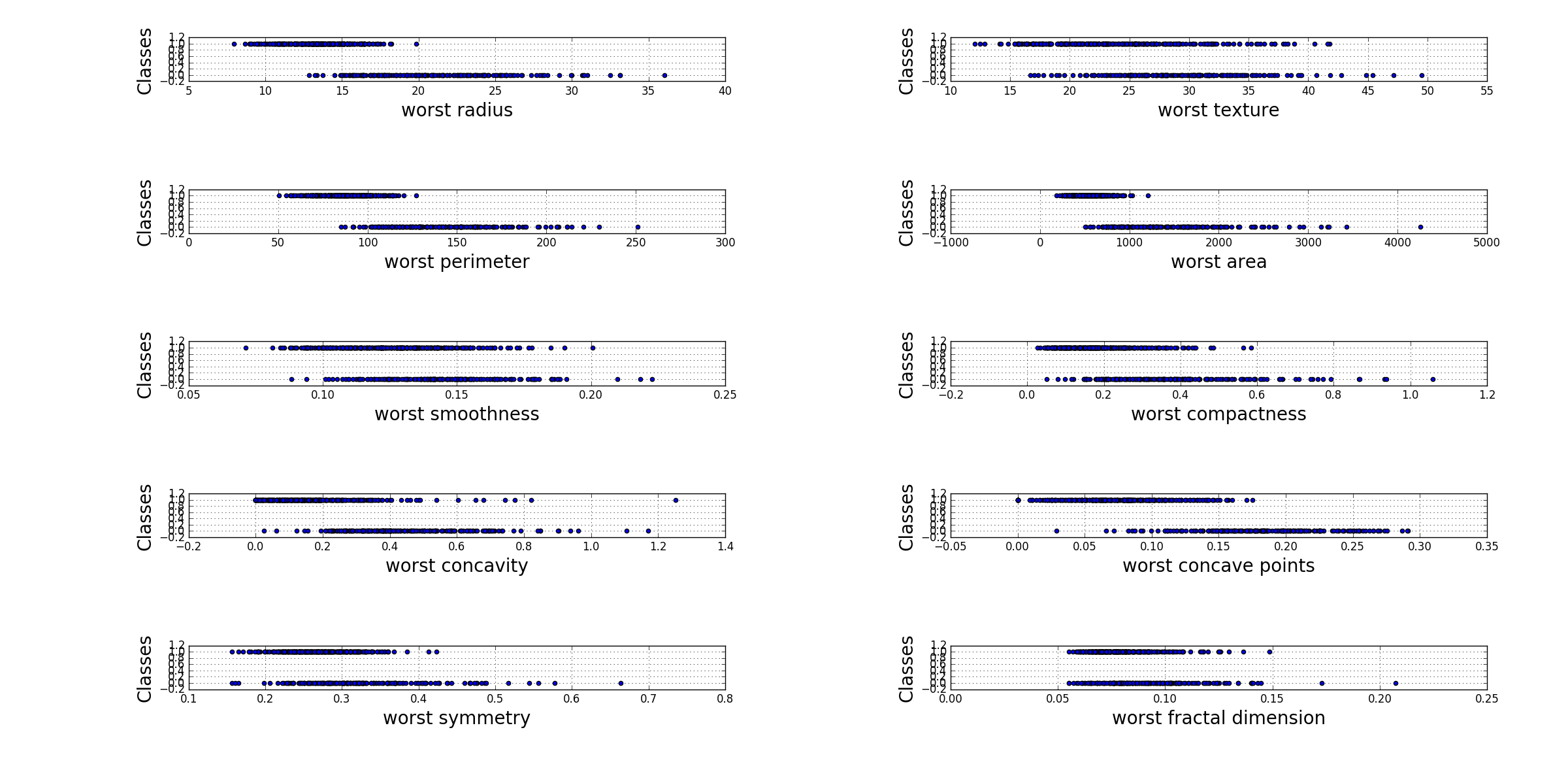}
	\caption{Plotted using \texttt{matplotlib}\cite{Hunter:2007}. Scatter plot of \textit{worst} features ($x_{20} - x_{29}$) in the WDBC.}
	\label{scatter-plot-worst}
\endminipage\hfill
\end{figure}
\indent	Figure \ref{training-accuracy} shows the training accuracy of the ML algorithms: (1) GRU-SVM finished its training in 2 minutes and 54 seconds with an average training accuracy of 90.6857639\%, (2) Linear Regression finished its training in 35 seconds with an average training accuracy of 92.8906257\%, (3) MLP finished its training in 28 seconds with an average training accuracy of 96.9286785\%, (4) Softmax Regression finished its training in 25 seconds with an average training accuracy of 97.366573\%, and (5) L2-SVM finished its training in 14 seconds with an average training accuracy of 97.734375\%. There was no recorded training accuracy for Nearest Neighbor search since it does not require any training, as the norm equations (Eq. \ref{taxicab-norm} and Eq. \ref{euclidean-norm}) are directly applied on the dataset to determine the ``nearest neighbor'' of a given data point $p_{i} \in \vec{p}$.\\
\indent	The empirical evidence presented in this section draws a qualitative comparability with, and corroborates the findings of \cite{zafiropoulos2006support}. Hence, a testament to the effectiveness of ML algorithms on the diagnosis of breast cancer. While the experiment results are all commendable, the performance of the GRU-SVM model\cite{agarap2017neural} warrants a discussion. The mid-level performance of GRU-SVM with a test accuracy of 93.75\% is hypothetically attributed to the following information: (1) the non-linearities introduced by the GRU model\cite{Cho} through its gating mechanism (see Eq. \ref{z-gate}, Eq. \ref{r-gate}, and Eq. \ref{candidate-value}) to its output may be the cause of a difficulty in generalizing on a linearly-separable data such as the WDBC dataset, and (2) the sensitivity of RNNs to weight initialization\cite{alalshekmubarak2013novel}. Since the weights of the GRU-SVM model are assigned with arbitrary values, it will also prove limited capability of result reproducibility, even when using an identical configuration\cite{alalshekmubarak2013novel}.\\
\indent	Despite the given arguments, it does not necessarily revoke the fact that GRU-SVM is comparable with the presented ML algorithms, as what the results have shown. In addition, it was a expected that the upper hand goes to the linear classifiers (Linear Regression and SVM) as the utilized dataset was linearly separable. The linear separability of the WDBC dataset is shown in a naive method of visualization (see Figure \ref{scatter-plot-mean}, Figure \ref{scatter-plot-error}, and Figure \ref{scatter-plot-worst}). Visually speaking, it is palpable that the scattered features in the mentioned figures may be easily separated by a linear function.

\section{Conclusion and Recommendation}

This paper presents an application of different machine learning algorithms, including the proposed GRU-SVM model in \cite{agarap2017neural}, for the diagnosis of breast cancer. All presented ML algorithms exhibited high performance on the binary classification of breast cancer, i.e. determining whether benign tumor or malignant tumor. Consequently, the statistical measures on the classification problem were also satisfactory.\\
\indent	To further substantiate the results of this study, a CV technique such as $k$-fold cross validation should be employed. The application of such a technique will not only provide a more accurate measure of model prediction performance, but it will also assist in determining the most optimal hyper-parameters for the ML algorithms\cite{bengio2015deep}.

\section{Acknowledgment}

% An expression of gratitude to Dr. William H. Wolberg of the University of Wisconsin for the WDBC dataset used in this study. Also, an appreciation of the open source community, especially Cross Validated, GitHub, Google, Python, and Stack Overflow.\\
\indent	Deep appreciation is given to the family and friends of the author (in arbitrary order): Myra M. Maranan, Faisal E. Montilla, Corazon Fabreag-Agarap, Crystal Love Fabreag-Agarap, Michaelangelo Milo L. Lim, Liberato F. Ramos, Hyacinth Gasmin, Rhea Jude Ferrer, Ma. Pauline de Ocampo, and Abqary Alon.

\bibliographystyle{ACM-Reference-Format}
\bibliography{paper} 


\begin{thebibliography}{21}

%%% the Web address.

\ifx \showCODEN    \undefined \def \showCODEN     #1{\unskip}     \fi
\ifx \showDOI      \undefined \def \showDOI       #1{#1}\fi
\ifx \showISBNx    \undefined \def \showISBNx     #1{\unskip}     \fi
\ifx \showISBNxiii \undefined \def \showISBNxiii  #1{\unskip}     \fi
\ifx \showISSN     \undefined \def \showISSN      #1{\unskip}     \fi
\ifx \showLCCN     \undefined \def \showLCCN      #1{\unskip}     \fi
\ifx \shownote     \undefined \def \shownote      #1{#1}          \fi
\ifx \showarticletitle \undefined \def \showarticletitle #1{#1}   \fi
\ifx \showURL      \undefined \def \showURL       {\relax}        \fi
% invisible to TeX
\providecommand\bibfield[2]{#2}
\providecommand\bibinfo[2]{#2}
\providecommand\natexlab[1]{#1}
\providecommand\showeprint[2][]{arXiv:#2}

\bibitem[\protect\citeauthoryear{??}{sur}{[n. d.]}]%
        {surveillance}
 \bibinfo{year}{[n. d.]}\natexlab{}.
\newblock   (\bibinfo{year}{[n. d.]}).
\newblock
\showURL{%
\url{https://seer.cancer.gov/statfacts/html/breast.html}}


\bibitem[\protect\citeauthoryear{??}{nci}{2017}]%
        {nci2017}
 \bibinfo{year}{2017}\natexlab{}.
\newblock \bibinfo{title}{Cancer Statistics}.
\newblock   (\bibinfo{date}{Mar} \bibinfo{year}{2017}).
\newblock
\showURL{%
\url{https://www.cancer.gov/about-cancer/understanding/statistics}}


\bibitem[\protect\citeauthoryear{Abadi, Agarwal, Barham, Brevdo, Chen, Citro,
  Corrado, Davis, Dean, Devin, Ghemawat, Goodfellow, Harp, Irving, Isard, Jia,
  Jozefowicz, Kaiser, Kudlur, Levenberg, Man\'{e}, Monga, Moore, Murray, Olah,
  Schuster, Shlens, Steiner, Sutskever, Talwar, Tucker, Vanhoucke, Vasudevan,
  Vi\'{e}gas, Vinyals, Warden, Wattenberg, Wicke, Yu, and Zheng}{Abadi
  et~al\mbox{.}}{2015}]%
        {tensorflow2015-whitepaper}
\bibfield{author}{\bibinfo{person}{Mart\'{\i}n Abadi}, \bibinfo{person}{Ashish
  Agarwal}, \bibinfo{person}{Paul Barham}, \bibinfo{person}{Eugene Brevdo},
  \bibinfo{person}{Zhifeng Chen}, \bibinfo{person}{Craig Citro},
  \bibinfo{person}{Greg~S. Corrado}, \bibinfo{person}{Andy Davis},
  \bibinfo{person}{Jeffrey Dean}, \bibinfo{person}{Matthieu Devin},
  \bibinfo{person}{Sanjay Ghemawat}, \bibinfo{person}{Ian Goodfellow},
  \bibinfo{person}{Andrew Harp}, \bibinfo{person}{Geoffrey Irving},
  \bibinfo{person}{Michael Isard}, \bibinfo{person}{Yangqing Jia},
  \bibinfo{person}{Rafal Jozefowicz}, \bibinfo{person}{Lukasz Kaiser},
  \bibinfo{person}{Manjunath Kudlur}, \bibinfo{person}{Josh Levenberg},
  \bibinfo{person}{Dan Man\'{e}}, \bibinfo{person}{Rajat Monga},
  \bibinfo{person}{Sherry Moore}, \bibinfo{person}{Derek Murray},
  \bibinfo{person}{Chris Olah}, \bibinfo{person}{Mike Schuster},
  \bibinfo{person}{Jonathon Shlens}, \bibinfo{person}{Benoit Steiner},
  \bibinfo{person}{Ilya Sutskever}, \bibinfo{person}{Kunal Talwar},
  \bibinfo{person}{Paul Tucker}, \bibinfo{person}{Vincent Vanhoucke},
  \bibinfo{person}{Vijay Vasudevan}, \bibinfo{person}{Fernanda Vi\'{e}gas},
  \bibinfo{person}{Oriol Vinyals}, \bibinfo{person}{Pete Warden},
  \bibinfo{person}{Martin Wattenberg}, \bibinfo{person}{Martin Wicke},
  \bibinfo{person}{Yuan Yu}, {and} \bibinfo{person}{Xiaoqiang Zheng}.}
  \bibinfo{year}{2015}\natexlab{}.
\newblock \bibinfo{title}{{TensorFlow}: Large-Scale Machine Learning on
  Heterogeneous Systems}.
\newblock   (\bibinfo{year}{2015}).
\newblock
\showURL{%
\url{http://tensorflow.org/}}
\newblock
\shownote{Software available from tensorflow.org.}


\bibitem[\protect\citeauthoryear{Agarap}{Agarap}{2017}]%
        {agarap2017neural}
\bibfield{author}{\bibinfo{person}{Abien~Fred Agarap}.}
  \bibinfo{year}{2017}\natexlab{}.
\newblock \showarticletitle{A Neural Network Architecture Combining Gated
  Recurrent Unit (GRU) and Support Vector Machine (SVM) for Intrusion Detection
  in Network Traffic Data}.
\newblock \bibinfo{journal}{{\em arXiv preprint arXiv:1709.03082\/}}
  (\bibinfo{year}{2017}).
\newblock


\bibitem[\protect\citeauthoryear{Alalshekmubarak and Smith}{Alalshekmubarak and
  Smith}{2013}]%
        {alalshekmubarak2013novel}
\bibfield{author}{\bibinfo{person}{Abdulrahman Alalshekmubarak} {and}
  \bibinfo{person}{Leslie~S Smith}.} \bibinfo{year}{2013}\natexlab{}.
\newblock \showarticletitle{A novel approach combining recurrent neural network
  and support vector machines for time series classification}. In
  \bibinfo{booktitle}{{\em Innovations in Information Technology (IIT), 2013
  9th International Conference on}}. IEEE, \bibinfo{pages}{42--47}.
\newblock


\bibitem[\protect\citeauthoryear{Bengio, Goodfellow, and Courville}{Bengio
  et~al\mbox{.}}{2015}]%
        {bengio2015deep}
\bibfield{author}{\bibinfo{person}{Yoshua Bengio}, \bibinfo{person}{Ian~J
  Goodfellow}, {and} \bibinfo{person}{Aaron Courville}.}
  \bibinfo{year}{2015}\natexlab{}.
\newblock \showarticletitle{Deep learning}.
\newblock \bibinfo{journal}{{\em Nature\/}}  \bibinfo{volume}{521}
  (\bibinfo{year}{2015}), \bibinfo{pages}{436--444}.
\newblock


\bibitem[\protect\citeauthoryear{Bishop}{Bishop}{1995}]%
        {bishop1995neural}
\bibfield{author}{\bibinfo{person}{Christopher~M Bishop}.}
  \bibinfo{year}{1995}\natexlab{}.
\newblock \bibinfo{booktitle}{{\em Neural networks for pattern recognition}}.
\newblock \bibinfo{publisher}{Oxford university press}.
\newblock


\bibitem[\protect\citeauthoryear{Cho, Van~Merri{\"e}nboer, Gulcehre, Bahdanau,
  Bougares, Schwenk, and Bengio}{Cho et~al\mbox{.}}{2014}]%
        {Cho}
\bibfield{author}{\bibinfo{person}{Kyunghyun Cho}, \bibinfo{person}{Bart
  Van~Merri{\"e}nboer}, \bibinfo{person}{Caglar Gulcehre},
  \bibinfo{person}{Dzmitry Bahdanau}, \bibinfo{person}{Fethi Bougares},
  \bibinfo{person}{Holger Schwenk}, {and} \bibinfo{person}{Yoshua Bengio}.}
  \bibinfo{year}{2014}\natexlab{}.
\newblock \showarticletitle{Learning phrase representations using RNN
  encoder-decoder for statistical machine translation}.
\newblock \bibinfo{journal}{{\em arXiv preprint arXiv:1406.1078\/}}
  (\bibinfo{year}{2014}).
\newblock


\bibitem[\protect\citeauthoryear{Cortes and Vapnik}{Cortes and Vapnik}{1995}]%
        {Cortes}
\bibfield{author}{\bibinfo{person}{C. Cortes} {and} \bibinfo{person}{V.
  Vapnik}.} \bibinfo{year}{1995}\natexlab{}.
\newblock \showarticletitle{Support-vector Networks}.
\newblock \bibinfo{journal}{{\em Machine Learning\/}}  \bibinfo{volume}{20.3}
  (\bibinfo{year}{1995}), \bibinfo{pages}{273--297}.
\newblock
\showDOI{%
\url{https://doi.org/10.1007/BF00994018}}


\bibitem[\protect\citeauthoryear{Damien}{Damien}{t 29}]%
        {aymericdamien}
\bibfield{author}{\bibinfo{person}{Aymeric Damien}.} \bibinfo{year}{2017,
  August 29}\natexlab{}.
\newblock   (\bibinfo{year}{2017, August 29}).
\newblock
\showURL{%
\url{https://github.com/aymericdamien/TensorFlow-Examples/blob/master/examples/2_BasicModels/nearest_neighbor.py}}
\newblock
\shownote{Accessed: November 17, 2017.}


\bibitem[\protect\citeauthoryear{Hahnloser, Sarpeshkar, Mahowald, Douglas, and
  Seung}{Hahnloser et~al\mbox{.}}{2000}]%
        {hahnloser2000digital}
\bibfield{author}{\bibinfo{person}{Richard~HR Hahnloser},
  \bibinfo{person}{Rahul Sarpeshkar}, \bibinfo{person}{Misha~A Mahowald},
  \bibinfo{person}{Rodney~J Douglas}, {and} \bibinfo{person}{H~Sebastian
  Seung}.} \bibinfo{year}{2000}\natexlab{}.
\newblock \showarticletitle{Digital selection and analogue amplification
  coexist in a cortex-inspired silicon circuit}.
\newblock \bibinfo{journal}{{\em Nature\/}} \bibinfo{volume}{405},
  \bibinfo{number}{6789} (\bibinfo{year}{2000}), \bibinfo{pages}{947--951}.
\newblock


\bibitem[\protect\citeauthoryear{Hunter}{Hunter}{2007}]%
        {Hunter:2007}
\bibfield{author}{\bibinfo{person}{J.~D. Hunter}.}
  \bibinfo{year}{2007}\natexlab{}.
\newblock \showarticletitle{Matplotlib: A 2D graphics environment}.
\newblock \bibinfo{journal}{{\em Computing In Science \& Engineering\/}}
  \bibinfo{volume}{9}, \bibinfo{number}{3} (\bibinfo{year}{2007}),
  \bibinfo{pages}{90--95}.
\newblock
\showDOI{%
\url{https://doi.org/10.1109/MCSE.2007.55}}


\bibitem[\protect\citeauthoryear{Kingma and Ba}{Kingma and Ba}{2014}]%
        {Kingma}
\bibfield{author}{\bibinfo{person}{Diederik Kingma} {and}
  \bibinfo{person}{Jimmy Ba}.} \bibinfo{year}{2014}\natexlab{}.
\newblock \showarticletitle{Adam: A method for stochastic optimization}.
\newblock \bibinfo{journal}{{\em arXiv preprint arXiv:1412.6980\/}}
  (\bibinfo{year}{2014}).
\newblock


\bibitem[\protect\citeauthoryear{McCulloch and Pitts}{McCulloch and
  Pitts}{1943}]%
        {mcculloch1943logical}
\bibfield{author}{\bibinfo{person}{Warren~S McCulloch} {and}
  \bibinfo{person}{Walter Pitts}.} \bibinfo{year}{1943}\natexlab{}.
\newblock \showarticletitle{A logical calculus of the ideas immanent in nervous
  activity}.
\newblock \bibinfo{journal}{{\em The bulletin of mathematical biophysics\/}}
  \bibinfo{volume}{5}, \bibinfo{number}{4} (\bibinfo{year}{1943}),
  \bibinfo{pages}{115--133}.
\newblock


\bibitem[\protect\citeauthoryear{Pedregosa, Varoquaux, Gramfort, Michel,
  Thirion, Grisel, Blondel, Prettenhofer, Weiss, Dubourg, Vanderplas, Passos,
  Cournapeau, Brucher, Perrot, and Duchesnay}{Pedregosa et~al\mbox{.}}{2011}]%
        {scikit-learn}
\bibfield{author}{\bibinfo{person}{F. Pedregosa}, \bibinfo{person}{G.
  Varoquaux}, \bibinfo{person}{A. Gramfort}, \bibinfo{person}{V. Michel},
  \bibinfo{person}{B. Thirion}, \bibinfo{person}{O. Grisel},
  \bibinfo{person}{M. Blondel}, \bibinfo{person}{P. Prettenhofer},
  \bibinfo{person}{R. Weiss}, \bibinfo{person}{V. Dubourg}, \bibinfo{person}{J.
  Vanderplas}, \bibinfo{person}{A. Passos}, \bibinfo{person}{D. Cournapeau},
  \bibinfo{person}{M. Brucher}, \bibinfo{person}{M. Perrot}, {and}
  \bibinfo{person}{E. Duchesnay}.} \bibinfo{year}{2011}\natexlab{}.
\newblock \showarticletitle{Scikit-learn: Machine Learning in {P}ython}.
\newblock \bibinfo{journal}{{\em Journal of Machine Learning Research\/}}
  \bibinfo{volume}{12} (\bibinfo{year}{2011}), \bibinfo{pages}{2825--2830}.
\newblock


\bibitem[\protect\citeauthoryear{Rosenblatt}{Rosenblatt}{1958}]%
        {rosenblatt1958perceptron}
\bibfield{author}{\bibinfo{person}{Frank Rosenblatt}.}
  \bibinfo{year}{1958}\natexlab{}.
\newblock \showarticletitle{The perceptron: A probabilistic model for
  information storage and organization in the brain.}
\newblock \bibinfo{journal}{{\em Psychological review\/}} \bibinfo{volume}{65},
  \bibinfo{number}{6} (\bibinfo{year}{1958}), \bibinfo{pages}{386}.
\newblock


\bibitem[\protect\citeauthoryear{Salama, Abdelhalim, and Zeid}{Salama
  et~al\mbox{.}}{2012}]%
        {salama2012breast}
\bibfield{author}{\bibinfo{person}{Gouda~I Salama}, \bibinfo{person}{M
  Abdelhalim}, {and} \bibinfo{person}{Magdy Abd-elghany Zeid}.}
  \bibinfo{year}{2012}\natexlab{}.
\newblock \showarticletitle{Breast cancer diagnosis on three different datasets
  using multi-classifiers}.
\newblock \bibinfo{journal}{{\em Breast Cancer (WDBC)\/}} \bibinfo{volume}{32},
  \bibinfo{number}{569} (\bibinfo{year}{2012}), \bibinfo{pages}{2}.
\newblock


\bibitem[\protect\citeauthoryear{Tang}{Tang}{2013}]%
        {Tang}
\bibfield{author}{\bibinfo{person}{Yichuan Tang}.}
  \bibinfo{year}{2013}\natexlab{}.
\newblock \showarticletitle{Deep learning using linear support vector
  machines}.
\newblock \bibinfo{journal}{{\em arXiv preprint arXiv:1306.0239\/}}
  (\bibinfo{year}{2013}).
\newblock


\bibitem[\protect\citeauthoryear{Walt, Colbert, and Varoquaux}{Walt
  et~al\mbox{.}}{2011}]%
        {walt2011numpy}
\bibfield{author}{\bibinfo{person}{St{\'e}fan van~der Walt},
  \bibinfo{person}{S~Chris Colbert}, {and} \bibinfo{person}{Gael Varoquaux}.}
  \bibinfo{year}{2011}\natexlab{}.
\newblock \showarticletitle{The NumPy array: a structure for efficient
  numerical computation}.
\newblock \bibinfo{journal}{{\em Computing in Science \& Engineering\/}}
  \bibinfo{volume}{13}, \bibinfo{number}{2} (\bibinfo{year}{2011}),
  \bibinfo{pages}{22--30}.
\newblock


\bibitem[\protect\citeauthoryear{Wolberg, Street, and Mangasarian}{Wolberg
  et~al\mbox{.}}{1992}]%
        {wolberg1992breast}
\bibfield{author}{\bibinfo{person}{William~H Wolberg}, \bibinfo{person}{W~Nick
  Street}, {and} \bibinfo{person}{Olvi~L Mangasarian}.}
  \bibinfo{year}{1992}\natexlab{}.
\newblock \showarticletitle{Breast cancer Wisconsin (diagnostic) data set}.
\newblock \bibinfo{journal}{{\em UCI Machine Learning Repository
  [http://archive. ics. uci. edu/ml/]\/}} (\bibinfo{year}{1992}).
\newblock


\bibitem[\protect\citeauthoryear{Zafiropoulos, Maglogiannis, and
  Anagnostopoulos}{Zafiropoulos et~al\mbox{.}}{2006}]%
        {zafiropoulos2006support}
\bibfield{author}{\bibinfo{person}{Elias Zafiropoulos}, \bibinfo{person}{Ilias
  Maglogiannis}, {and} \bibinfo{person}{Ioannis Anagnostopoulos}.}
  \bibinfo{year}{2006}\natexlab{}.
\newblock \showarticletitle{A support vector machine approach to breast cancer
  diagnosis and prognosis}.
\newblock \bibinfo{journal}{{\em Artificial Intelligence Applications and
  Innovations\/}} (\bibinfo{year}{2006}), \bibinfo{pages}{500--507}.
\newblock


\end{thebibliography}

\end{document}